\pdfoutput=1
\documentclass[twocolumn]{svjour3}

\smartqed  

\usepackage{graphicx} 
\usepackage{subfigure}
\usepackage{float} 
\usepackage{wrapfig} 
\usepackage{amsmath,amssymb} 
\usepackage{shorttoc}
\usepackage{multirow}
\usepackage{booktabs}
\usepackage{algorithm}
\usepackage{algorithmic}
\usepackage{setspace}
\usepackage{epsfig}
\usepackage{epstopdf}

\begin{document}

\title{A review of heterogeneous data mining for brain disorders}

\titlerunning{A review of heterogeneous data mining for brain disorders}

\author{
Bokai~Cao \and
Xiangnan~Kong \and
Philip~S.~Yu
}

\authorrunning{B. Cao et al.}

\institute{
B. Cao \and P.S. Yu \at
Department of Computer Science, University of Illinois at Chicago, Chicago, IL 60607. \\
\email{caobokai@uic.edu, psyu@cs.uic.edu} \and
X. Kong \at
Department of Computer Science, Worcester Polytechnic Institute, Worcester, MA 01609. \\
\email{xkong@wpi.edu}
}

\date{
}

\maketitle

\begin{abstract}
With rapid advances in neuroimaging techniques, the research on brain disorder identification has become an emerging area in the data mining community. Brain disorder data poses many unique challenges for data mining research. For example, the raw data generated by neuroimaging experiments is in tensor representations, with typical characteristics of high dimensionality, structural complexity and nonlinear separability. Furthermore, brain connectivity networks can be constructed from the tensor data, embedding subtle interactions between brain regions. Other clinical measures are usually available reflecting the disease status from different perspectives. It is expected that integrating complementary information in the tensor data and the brain network data, and incorporating other clinical parameters will be potentially transformative for investigating disease mechanisms and for informing therapeutic interventions. Many research efforts have been devoted to this area. They have achieved great success in various applications, such as tensor-based modeling, subgraph pattern mining, multi-view feature analysis. In this paper, we review some recent data mining methods that are used for analyzing brain disorders.
\keywords{Data mining \and Brain diseases \and Tensor analysis \and Subgraph patterns \and Feature selection}
\end{abstract}

\section{Introduction}\label{sec:introduction}

Many brain disorders are characterized by ongoing injury that is clinically silent for prolonged periods and irreversible by the time symptoms first present. New approaches for detection of early changes in subclinical periods will afford powerful tools for aiding clinical diagnosis, clarifying underlying mechanisms and informing neuroprotective interventions to slow or reverse neural injury for a broad spectrum of brain disorders, including bipolar disorder, HIV infection on brain, Alzheimer's disease, Parkinson's disease, \emph{etc.} Early diagnosis has the potential to greatly alleviate the burden of brain disorders and the ever increasing costs to families and society.

As the identification of brain disorders is extremely challenging, many different diagnosis tools and methods have been developed to obtain a large number of measurements from various examinations and laboratory tests. Especially, recent advances in the neuroimaging technology have provided an efficient and noninvasive way for studying the structural and functional connectivity of the human brain, either normal or in a diseased state \cite{rubinov2010complex}. This can be attributed in part to advances in magnetic resonance imaging (MRI) capabilities \cite{kong2014brain}.
Techniques such as diffusion MRI, also referred to as diffusion tensor imaging (DTI), produce in vivo images of the diffusion process of water molecules in biological tissues. By leveraging the fact that the water molecule diffusion patterns reveal microscopic details about tissue architecture, DTI can be used to perform tractography within the white matter and construct structural connectivity networks \cite{basser1996microstructural,le1986mr,chenevert1990anisotropic,mckeown1998analysis,moseley1990diffusion}.
Functional MRI (fMRI) is a functional neuroimaging procedure that identifies localized patterns of brain activation by detecting associated changes in the cerebral blood flow. The primary form of fMRI uses the blood oxygenation level dependent (BOLD) response extracted from the gray matter \cite{biswal1995functional,ogawa1990brain,ogawa1990oxygenation}. 
Another neuroimaging technique is positron emission tomography (PET). Using different radioactive tracers (\emph{e.g.}, fluorodeoxyglucose), PET produces a three-dimensional image of various physiological, biochemical and metabolic processes \cite{ye2008heterogeneous}.

A variety of data representations can be derived from these neuroimaging experiments, which present many unique challenges for the data mining community. Conventional data mining algorithms are usually developed to tackle data in one specific representation, a majority of which are particularly for vector-based data. However, the raw neuroimaging data is in the form of tensors, from which we can further construct brain networks connecting regions of interest (ROIs). Both of them are highly structured considering correlations between adjacent voxels in the tensor data and that between connected brain regions in the brain network data. Moreover, it is critical to explore interactions between measurements computed from the neuroimaging and other clinical experiments which describe subjects in different vector spaces. In this paper, we review some recent data mining methods for (1) mining tensor imaging data; (2) mining brain networks; (3) mining multi-view feature vectors.

\section{Tensor Imaging Analysis}


For brain disorder identification, the raw data generated by neuroimging experiments are in tensor representations \cite{davidson2013network,he3dusk,ye2008heterogeneous}. For example, in contrast to two-dimensional X-ray images, an fMRI sample corresponds to a four-dimensional array by recording the sequential changes of traceable signals in each voxel\footnote{A voxel is the smallest three-dimensional point volume referenced in a neuroimaging of the brain.}.

Tensors are higher order arrays that generalize the concepts of vectors (first-order tensors) and matrices (second-order tensors), whose elements are indexed by more than two indices. Each index expresses a \emph{mode} of variation of the data and corresponds to a coordinate direction. In an fMRI sample, the first three modes usually encode the spatial information, while the fourth mode encodes the temporal information. The number of variables in each mode indicates the dimensionality of a mode. The order of a tensor is determined by the number of its modes. An $m$th-order tensor can be represented as $\mathcal{X}=(x_{i_1,\cdots,i_m})\in\mathbb{R}^{I_{1}\times\cdots\times I_{m}}$, where $I_i$ is the dimension of $\mathcal{X}$ along the $i$-th mode.

\begin{definition}[Tensor product] The tensor product of three vectors $\mathbf{a} \in \mathbb{R}^{I_{1}}$, $\mathbf{b} \in \mathbb{R}^{I_{2}}$ and $\mathbf{c} \in \mathbb{R}^{I_{3}}$, denoted by $\mathbf{a} \otimes \mathbf{b} \otimes \mathbf{c}$, represents a third-order tensor with the elements $\left(\mathbf{a} \otimes \mathbf{b} \otimes \mathbf{c} \right)_{i_1, i_2, i_3}\ =\ a_{i_1}b_{i_2}c_{i_3}$.
\end{definition}

Tensor product is also referred to as outer product in some literature. An $m$th-order tensor is a rank-one tensor if it can be defined as the tensor product of $m$ vectors. 

\begin{figure}[t]
\centering
    \begin{minipage}[l]{\columnwidth}
      \centering
      \includegraphics[width=1\textwidth]{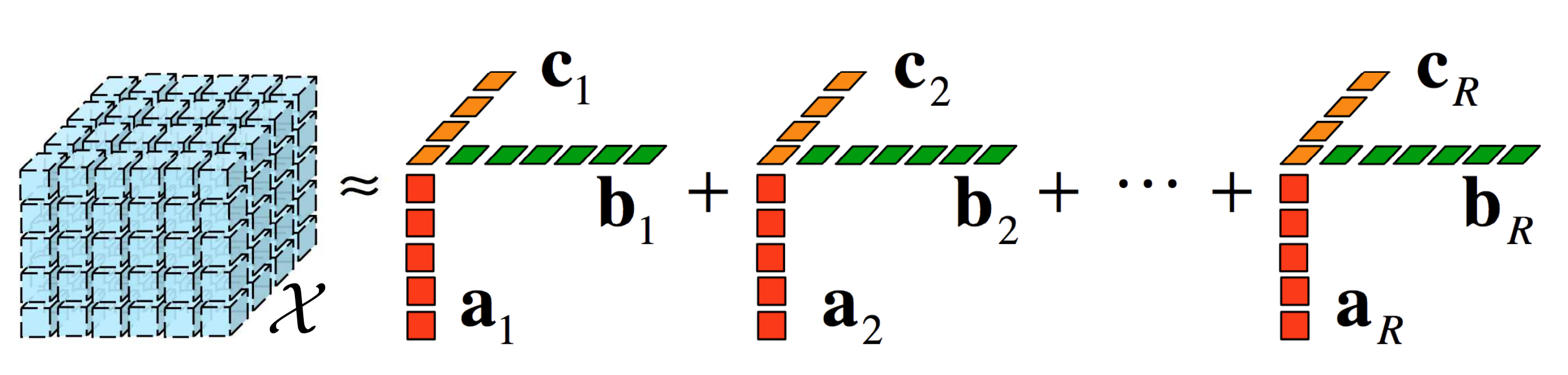}
    \end{minipage}
  \caption{Tensor factorization of a third-order tensor.}\label{fig:cp}
\end{figure}

\begin{definition}[Tensor factorization] Given a third-order tensor $\mathcal{X} \in \mathbb{R}^{I_{1} \times I_{2} \times I_{3}}$ and an integer $R$, as illustrated in Figure~\ref{fig:cp}, a tensor factorization of $\mathcal{X}$ can be expressed as
\begin{equation}\label{eq:cp}
\mathcal{X}=\mathcal{X}_1+\mathcal{X}_2+\cdots+\mathcal{X}_R=\sum_{r=1}^{R}\mathbf{a}_r\otimes\mathbf{b}_r\otimes\mathbf{c}_r
\end{equation}
\end{definition}

One of the major difficulties brought by the tensor data is the curse of dimensionality. The total number of voxels contained in a multi-mode tensor, say, $\mathcal{X}=(x_{i_1,\cdots,i_m})\in\mathbb{R}^{I_{1}\times\cdots\times I_{m}}$ is $I_{1}\times\cdots\times I_{m}$ which is exponential to the number of modes. If we unfold the tensor into a vector, the number of features will be extremely high \cite{zhou2013tensor}. This makes traditional data mining methods prone to overfitting, especially with a small sample size. Both computational scalability and theoretical guarantee of the traditional models are compromised by such high dimensionality \cite{he3dusk}.

On the other hand, complex structural information is embedded in the tensor data. For example, in the neuroimaging data, values of adjacent voxels are usually correlated with each other \cite{kong2014brain}. Such spatial relationships among different voxels in a tensor image can be very important in neuroimaging applications. Conventional tensor-based approaches focus on reshaping the tensor data into matrices/vectors and thus the original spatial relationships are lost. The integration of structural information is expected to improve the accuracy and interpretability of tensor models.

\subsection{Classification}
\label{sec:tensor_cla}

Suppose we have a set of tensor data $\mathcal{D}=\{(\mathcal{X}_i,y_i)\}_{i=1}^n$ for classification problem, where $\mathcal{X}_i \in \mathbb{R}^{I_{1} \times \cdots \times I_{m}}$ is the neuroimaging data represented as an $m$th-order tensor and $y_i\in\{-1,+1\}$ is the corresponding binary class label of $\mathcal{X}_i$. For example, if the $i$-th subject has Alzheimer's disease, the subject is associated with a positive label, \emph{i.e.}, $y_i=+1$. Otherwise, if the subject is in the control group, the subject is associated with a negative label, \emph{i.e.}, $y_i=-1$.

Supervised tensor learning can be formulated as the optimization problem of support tensor machines (STMs) \cite{tao2007supervised} which is a generalization of the standard support vector machines (SVMs) from vector data to tensor data. The objective of such learning algorithms is to learn a hyperplane by which the samples with different labels are divided as wide as possible. However, tensor data may not be linearly separable in the input space. To achieve a better performance on finding the most discriminative biomarkers or identifying infected subjects from the control group, in many neuroimaging applications, nonlinear transformation of the original tensor data should be considered. He et al. study the problem of supervised tensor learning with nonlinear kernels which can preserve the structure of tensor data \cite{he3dusk}. The proposed kernel is an extension of kernels in the vector space to the tensor space which can take the multidimensional structure complexity into account.

\subsection{Regression}
\label{sec:tensor_reg}

Slightly different from classifying disease status (discrete label), another family of problems use tensor neuroimages to predict cognitive outcome (continuous label). The problems can be formulated in a regression setup by treating clinical outcome as the real label, \emph{i.e.}, $y_i\in\mathbb{R}$, and treating tensor neuroimages as the input. However, most classical regression methods take vectors as input features. Simply reshaping a tensor into a vector is clearly an unsatisfactory solution.

Zhou et al. exploit the tensor structure in imaging data and integrate tensor decomposition within a statistical regression paradigm to model multidimensional arrays \cite{zhou2013tensor}. By imposing a low rank approximation to the extremely high dimensional complex imaging data, the curse of dimensionality is greatly alleviated, thereby allowing development of a fast estimation algorithm and regularization. Numerical analysis demonstrates its potential applications in identifying regions of interest in brains that are relevant to a particular clinical response.

\subsection{Network Discovery}
\label{sec:tensor_net}

Modern imaging techniques have allowed us to study the human brain as a complex system by modeling it as a network \cite{ajilore2013constructing}. For example, the fMRI scans consist of activations of thousands of voxels over time embedding a complex interaction of signals and noise \cite{genovese2002thresholding}, which naturally presents the problem of eliciting the underlying network from brain activities in the spatio-temporal tensor data. A brain connectivity network, also called a connectome \cite{sporns2005human}, consists of nodes (gray matter regions) and edges (white matter tracts in structural networks or correlations between two BOLD time series in functional networks).

Although the anatomical atlases in the brain have been extensively studied for decades, task/subject specific networks have still not been completely explored with consideration of functional or structural connectivity information. An anatomically parcellated region may contain subregions that are characterized by dramatically different functional or structural connectivity patterns, thereby significantly limiting the utility of the constructed networks. There are usually trade-offs between reducing noise and preserving utility in brain parcellation \cite{kong2014brain}. Thus investigating how to directly construct brain networks from tensor imaging data and understanding how they develop, deteriorate and vary across individuals will benefit disease diagnosis \cite{davidson2013network}.

Davidson et al. pose the problem of network discovery from fMRI data which involves simplifying spatio-temporal data into regions of the brain (nodes) and relationships between those regions (edges) \cite{davidson2013network}. Here the nodes represent collections of voxels that are known to behave cohesively over time; the edges can indicate a number of properties between nodes such as facilitation/inhibition (increases/decreases activity) or probabilistic (synchronized activity) relationships; the weight associated with each edge encodes the strength of the relationship.

A tensor can be decomposed into several factors. However, unconstrained tensor decomposition results of the fMRI data may not be good for node discovery because each factor is typically not a spatially contiguous region nor does it necessarily match an anatomical region. That is to say, many spatially adjacent voxels in the same structure are not active in the same factor which is anatomically impossible. Therefore, to achieve the purpose of discovering nodes while preserving anatomical adjacency, known anatomical regions in the brain are used as masks and constraints are added to enforce that the discovered factors should closely match these masks \cite{davidson2013network}.


Overall, current research on tensor imaging analysis presents two directions: (1) supervised: for a particular brain disorder, a classifier can be trained by modeling the relationship between a set of neuroimages and their associated labels (disease status or clinical response); (2) unsupervised: regardless of brain disorders, a brain network can be discovered from a given neuroimage.

\section{Brain Network Analysis}


We have briefly introduced that brain networks can be constructed from neuroimaging data where nodes correspond to brain regions, \emph{e.g.}, \emph{insula}, \emph{hippocampus}, \emph{thalamus}, and links correspond to the functional/structural connectivity between brain regions. The linkage structure in brain networks can encode tremendous information about the mental health of human subjects. For example, in brain networks derived from functional magnetic resonance imaging (fMRI), functional connections can encode the correlations between the functional activities of brain regions. While structural links in diffusion tensor imaging (DTI) brain networks can capture the number of neural fibers connecting different brain regions. The complex structures and the lack of vector representations for the brain network data raise major challenges for data mining.

Next, we will discuss different approaches on how to conduct further analysis for constructed brain networks, which are also referred to as graphs hereafter.

\begin{definition}[Binary graph] A binary graph is represented as $G =(V,E)$, where $V=\{v_1,\cdots,v_{n_v}\}$ is the set of vertices, $E\subseteq V\times V$ is the set of deterministic edges.
\end{definition}

\subsection{Kernel Learning on Graphs}

In the setting of supervised learning on graphs, the target is to train a classifier using a given set of graph data $\mathcal{D}=\{(G_i,y_i)\}_{i=1}^n$, so that we can predict the label $\hat{y}$ for a test graph $G$. With applications to brain networks, it is desirable to identify the disease status for a subject based on his/her uncovered brain network. Recent development of brain network analysis has made characterization of brain disorders at a whole-brain connectivity level possible, thus providing a new direction for brain disease classification.


Due to the complex structures and the lack of vector representations, graph data can not be directly used as the input for most data mining algorithms. A straightforward solution that has been extensively explored is to first derive features from brain networks and then construct a kernel on the feature vectors.

Wee et al. use brain connectivity networks for disease diagnosis on mild cognitive impairment (MCI), which is an early phase of Alzheimer's disease (AD) and usually regarded as a good target for early diagnosis and therapeutic interventions \cite{wee2012resting,wee2011enriched,wee2012identification}.
In the step of feature extraction, weighted local clustering coefficients of each ROI in relation to the remaining ROIs are extracted from all the constructed brain networks to quantify the prevalence of clustered connectivity around the ROIs.
To select the most discriminative features for classification, statistical t-test is performed and features with p-values smaller than a predefined threshold are selected to construct a kernel matrix. Through the employment of the multi-kernel SVM, Wee et al. integrate information from DTI and fMRI and achieve accurate early detection of brain abnormalities \cite{wee2012identification}.




However, such strategy simply treats a graph as a collection of nodes/links, and then extracts local measures (\emph{e.g.}, clustering coefficient) for each node or performs statistical analysis on each link, thereby blinding the connectivity structures of brain networks. Motivated by the fact that some data in real-world applications are naturally represented by means of graphs, while compressing and converting them to vectorial representations would definitely lose structural information, kernel methods for graphs have been extensively studied for a decade \cite{camastra2008kernel}.

A graph kernel maps the graph data from the original graph space to the feature space and further measures the similarity between two graphs by comparing their topological structures \cite{shervashidze2011weisfeiler}. For example, product graph kernel is based on the idea of counting the number of walks in product graphs \cite{gartner2003graph}; marginalized graph kernel works by comparing the label sequences generated by synchronized random walks of labeled graphs \cite{kashima2003marginalized}; cyclic pattern kernels for graphs count pairs of matching cyclic/tree patterns in two graphs \cite{horvath2004cyclic}. 

To identify individuals with AD/MCI from healthy controls, instead of using only a single property of brain networks, Jie et al. integrate multiple properties of fMRI brain networks to improve the disease diagnosis performance \cite{jie2014integration}.
Two different yet complementary network properties, \emph{i.e.}, local connectivity and global topological properties are quantified by computing two different types of kernels, \emph{i.e.}, a vector-based kernel and a graph kernel. As a local network property, weighted clustering coefficients are extracted to compute a vector-based kernel. As a topology-based graph kernel, Weisfeiler-Lehman subtree kernel \cite{shervashidze2011weisfeiler} is used to measure the topological similarity between paired fMRI brain networks. It is shown that this type of graph kernel can effectively capture the topological information from fMRI brain networks. The multi-kernel SVM is employed to fuse these two heterogeneous kernels for distinguishing individuals with MCI from healthy controls.

\subsection{Subgraph Pattern Mining}\label{sec:graph_pattern}

In brain network analysis, the ideal patterns we want to mine from the data should take care of both local and global graph topological information. Graph kernel methods seem promising, which however are not interpretable. Subgraph patterns are more suitable for brain networks, which can simultaneously model the network connectivity patterns around the nodes and capture the changes in local area \cite{kong2014brain}.

\begin{definition}[Subgraph]
Let $G'=(V',E')$ and $G=(V,E)$ be two binary graphs. $G'$ is a subgraph of $G$ (denoted as $G'\subseteq G$) iff $V'\subseteq V$ and $E'\subseteq E$. If $G'$ is a subgraph of $G$, then $G$ is supergraph of $G'$.
\end{definition}

\begin{figure}[t]
\centering
    \begin{minipage}[l]{\columnwidth}
      \centering
      \includegraphics[width=1\textwidth]{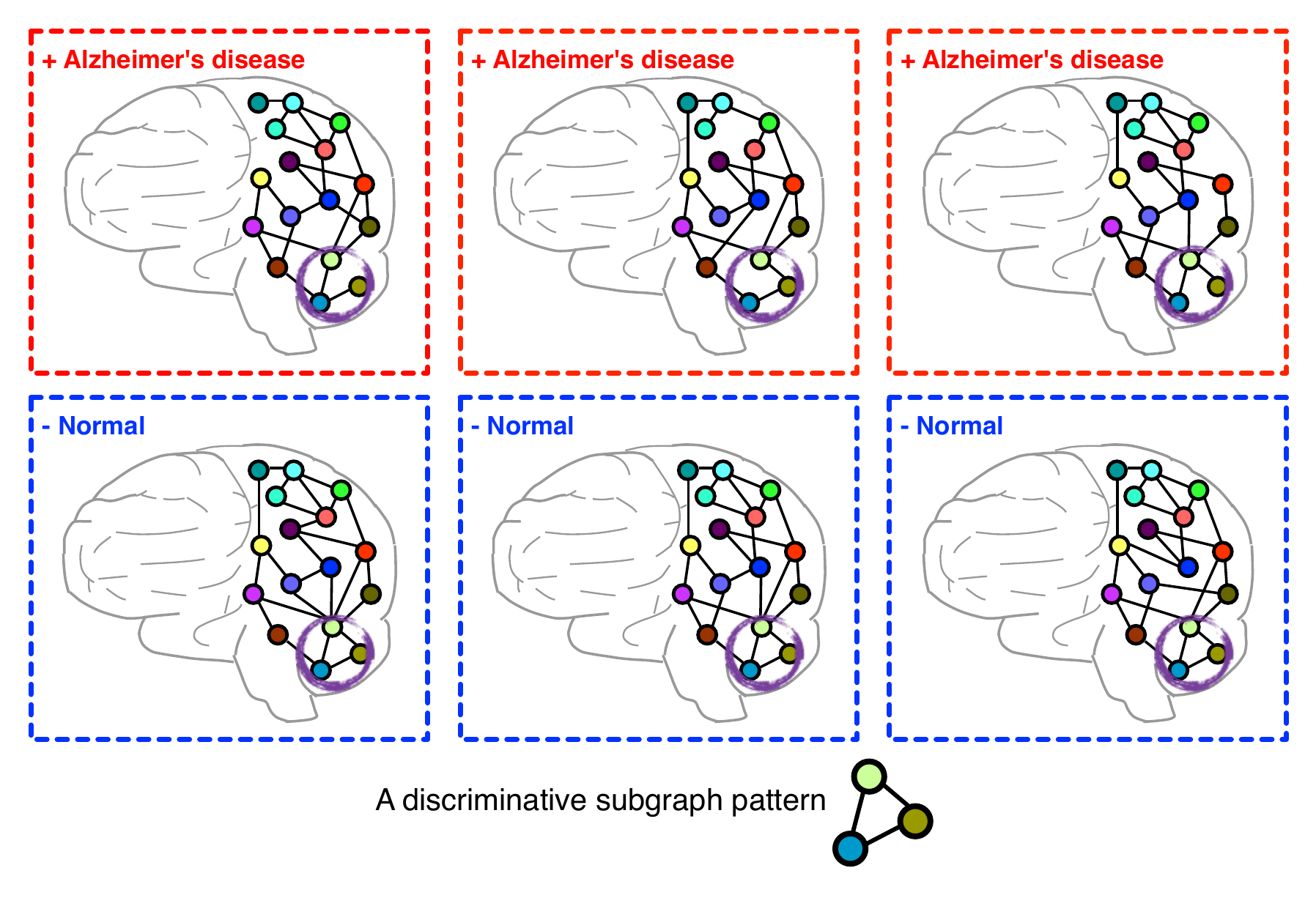}
    \end{minipage}
  \caption{An example of discriminative subgraph patterns in brain networks.}\label{fig:subgraph}
\end{figure}

A subgraph pattern, in a brain network, represents a collection of brain regions and their connections. For example, as shown in Figure~\ref{fig:subgraph}, three brain regions should work collaboratively for normal people and the absence of any connection between them can result in Alzheimer's disease in different degree. Therefore, it is valuable to understand which connections collectively play a significant role in disease mechanism by finding discriminative subgraph patterns in brain networks.

Mining subgraph patterns from graph data has been extensively studied by many researchers \cite{jin2010gaia,cheng2009identifying,thoma2009near,yan2008mining}. In general, a variety of filtering criteria are proposed. A typical evaluation criterion is frequency, which aims at searching for frequently appearing subgraph features in a graph dataset satisfying a prespecified threshold. Most of the frequent subgraph mining approaches are unsupervised. For example, Yan and Han develop a depth-first search algorithm: gSpan \cite{yan2002gspan}. This algorithm builds a lexicographic order among graphs, and maps each graph to an unique minimum DFS code as its canonical label. Based on this lexicographic order, gSpan adopts the depth-first search strategy to mine frequent connected subgraphs efficiently. Many other approaches for frequent subgraph mining have also been proposed, \emph{e.g.}, AGM \cite{inokuchi2000apriori}, FSG \cite{kuramochi2001frequent}, MoFa \cite{borgelt2002mining}, FFSM \cite{huan2003efficient}, and Gaston \cite{nijssen2004quickstart}.

Moreover, the problem of supervised subgraph mining has been studied in recent work which examines how to improve the efficiency of searching the discriminative subgraph patterns for graph classification. Yan et al. introduce two concepts \emph{structural leap search} and \emph{frequency-descending mining}, and propose LEAP \cite{yan2008mining} which is one of the first work in discriminative subgraph mining. Thoma et al. propose CORK which can yield a near-optimal solution using greedy feature selection \cite{thoma2009near}. Ranu and Singh propose a scalable approach, called GraphSig, that is capable of mining discriminative subgraphs with a low frequency threshold \cite{ranu2009graphsig}. Jin et al. propose COM which takes into account the co-occurences of subgraph patterns, thereby facilitating the mining process \cite{jin2009graph}. Jin et al. further propose an evolutionary computation method, called GAIA, to mine discriminative subgraph patterns using a randomized searching strategy \cite{jin2010gaia}. Zhu et al. design a diversified discrimination score based on the log ratio which can reduce the overlap between selected features by considering the embedding overlaps in the graphs \cite{zhu2012graph}.

\begin{figure}[t]
\centering
    \begin{minipage}[l]{\columnwidth}
      \centering
      \includegraphics[width=1\textwidth]{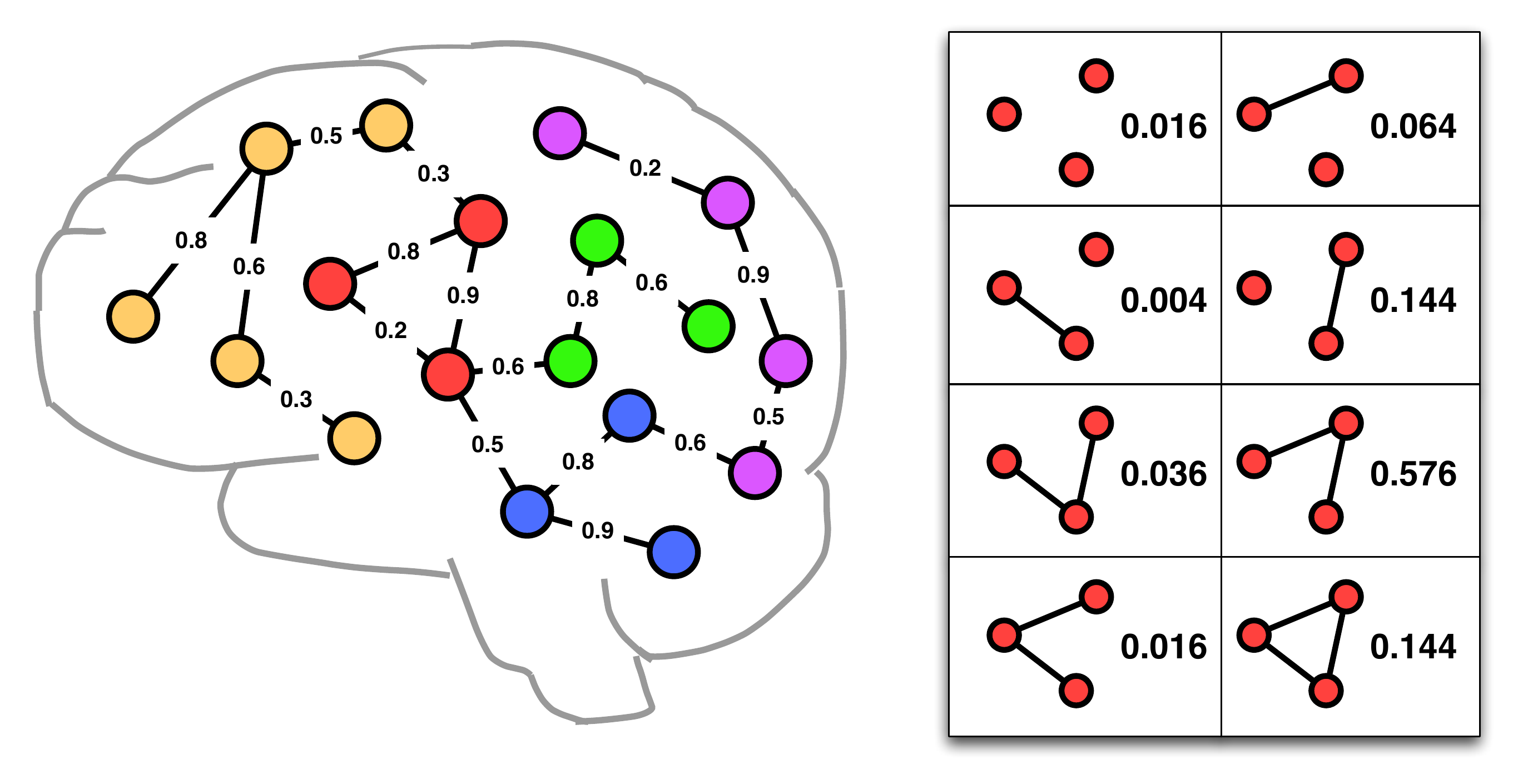}
    \end{minipage}
  \caption{An example of fMRI brain networks (left) and all possible instantiations of linkage structures between red nodes (right) \cite{cao2015identification}.}\label{fig:uncertain}
\end{figure}

Conventional graph mining approaches are best suited for binary edges, where the structure of graph objects is deterministic, and the binary edges represent the presence of linkages between the nodes \cite{kong2014brain}. In fMRI brain network data however, there are inherently weighted edges in the graph linkage structure, as shown in Figure~\ref{fig:uncertain} (left). A straightforward solution is to threshold weighted networks to yield binary networks. However, such simplification will result in great loss of information. Ideal data mining methods for brain network analysis should be able to overcome these methodological problems by generalizing the network edges to positive and negative weighted cases, \emph{e.g.}, probabilistic weights in fMRI brain networks, integral weights in DTI brain networks.

\begin{definition}[Weighted graph] A weighted graph is represented as $\widetilde{G}=(V,E,p)$, where $V=\{v_1,\cdots,v_{n_v}\}$ is the set of vertices, $E\subseteq V\times V$ is the set of nondeterministic edges. $p:E\rightarrow(0,1]$ is a function that assigns a probability of existence to each edge in $E$.
\end{definition}

fMRI brain networks can be modeled as weighted graphs where each edge $e\in E$ is associated with a probability $p(e)$ indicating the likelihood of whether this edge should exist or not \cite{kong2013discriminative,cao2015identification}. It is assumed that $p(e)$ of different edges in a weighted graph are independent from each other. Therefore, by enumerating the possible existence of all edges in a weighted graph, we can obtain a set of binary graphs. For example, in Fig.~\ref{fig:uncertain} (right), consider the three red nodes and links between them as a weighted graph. There are $2^3=8$ binary graphs that can be implied with different probabilities. For a weighted graph $\widetilde{G}$, the probability of $\widetilde{G}$ containing a subgraph feature $G'$ is defined as the probability that a binary graph $G$ implied by $\widetilde{G}$ contains subgraph $G'$. Kong et al. propose a discriminative subgraph feature selection method based on dynamic programming to compute the probability distribution of the discrimination scores for each subgraph pattern within a set of weighted graphs \cite{kong2013discriminative}.



For brain network analysis, usually we only have a small number of graph instances \cite{kong2013discriminative}. In these applications, the graph view alone is not sufficient for mining important subgraphs. Fortunately, the side information is available along with the graph data for brain disorder identification. For example, in neurological studies, hundreds of clinical, immunologic, serologic and cognitive measures may be available for each subject, apart from brain networks. These measures compose multiple side views which contain a tremendous amount of supplemental information for diagnostic purposes. It is desirable to extract valuable information from a plurality of side views to guide the process of subgraph mining in brain networks.

\begin{figure}[t]
\centering
\subfigure[Treating side views and subgraph patterns separately.]{\label{fig:side_method1}
    \begin{minipage}[l]{\columnwidth}
      \centering
      \includegraphics[width=1\textwidth]{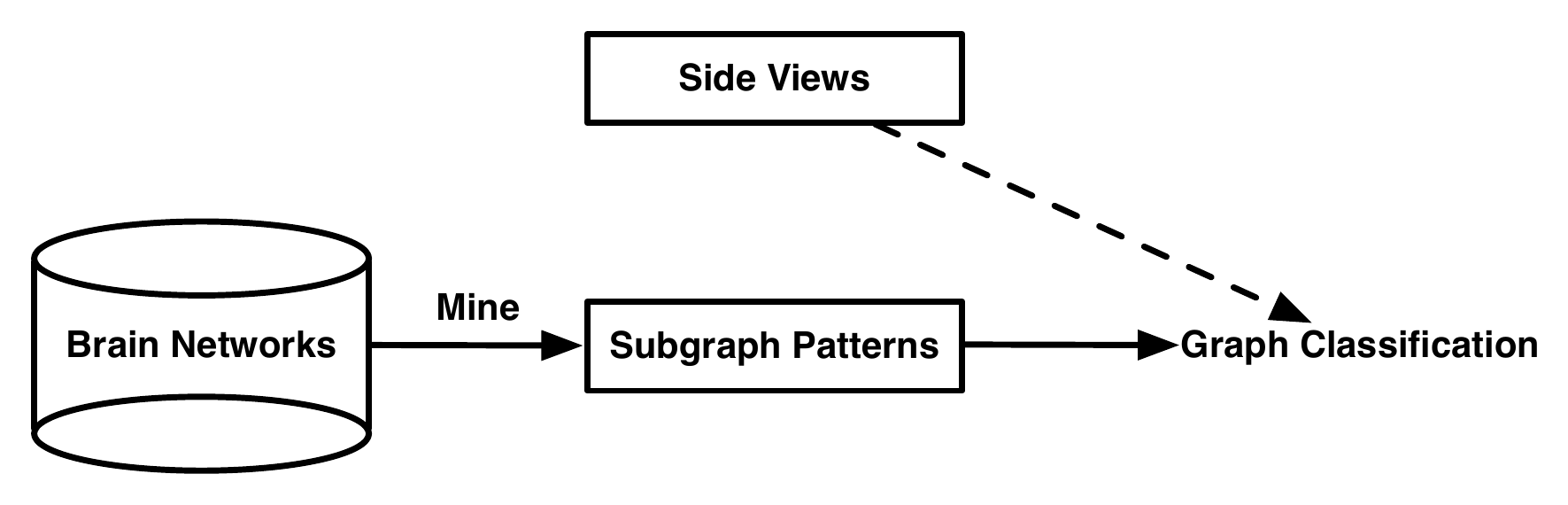}
    \end{minipage}
  }
\subfigure[Using side views as guidance for the process of selecting subgraph patterns.]{\label{fig:side_method2}
    \begin{minipage}[l]{\columnwidth}
      \centering
      \includegraphics[width=1\textwidth]{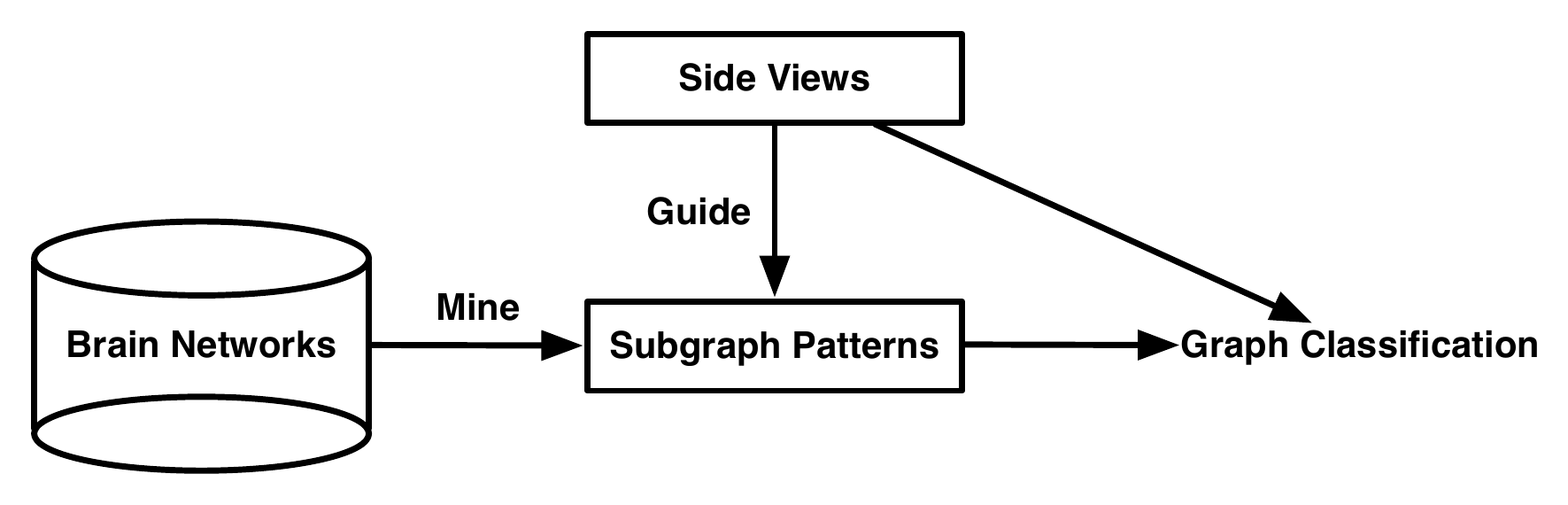}
    \end{minipage}
  }
\caption{Two strategies of leveraging side views in feature selection process for graph classification.}\label{fig:side_method}
\end{figure}

Figure~\ref{fig:side_method1} illustrates the process of selecting subgraph patterns in conventional graph classification approaches. Obviously, the valuable information embedded in side views is not fully leveraged in feature selection process.
To tackle this problem, Cao et al. introduce an effective algorithm for discriminative subgraph selection using multiple side views \cite{cao2015mining}, as illustrated in Figure~\ref{fig:side_method2}. Side information consistency is first validated via statistical hypothesis testing which suggests that the similarity of side view features between instances with the same label should have higher probability to be larger than that with different labels. Based on such observations, it is assumed that the similarity/distance between instances in the space of subgraph features should be consistent with that in the space of a side view. That is to say, if two instances are similar in the space of a side view, they should also be close to each other in the space of subgraph features. Therefore the target is to minimize the distance between subgraph features of each pair of similar instances in each side view \cite{cao2015mining}. In contrast to existing subgraph mining approaches that focus on the graph view alone, the proposed method can explore multiple vector-based side views to find an optimal set of subgraph features for graph classification.

For graph classification, brain network analysis approaches can generally be put into three groups: (1) extracting some local measures (\emph{e.g.}, clustering coefficient) to train a standard vector-based classifier; (2) directly adopting graph kernels for classification; (3) finding discriminative subgraph patterns. Different types of methods model the connectivity embedded in brain networks in different ways.

\section{Multi-view Feature Analysis}

\begin{figure}[t]
\centering
    \begin{minipage}[l]{\columnwidth}
      \centering
      \includegraphics[width=1\textwidth]{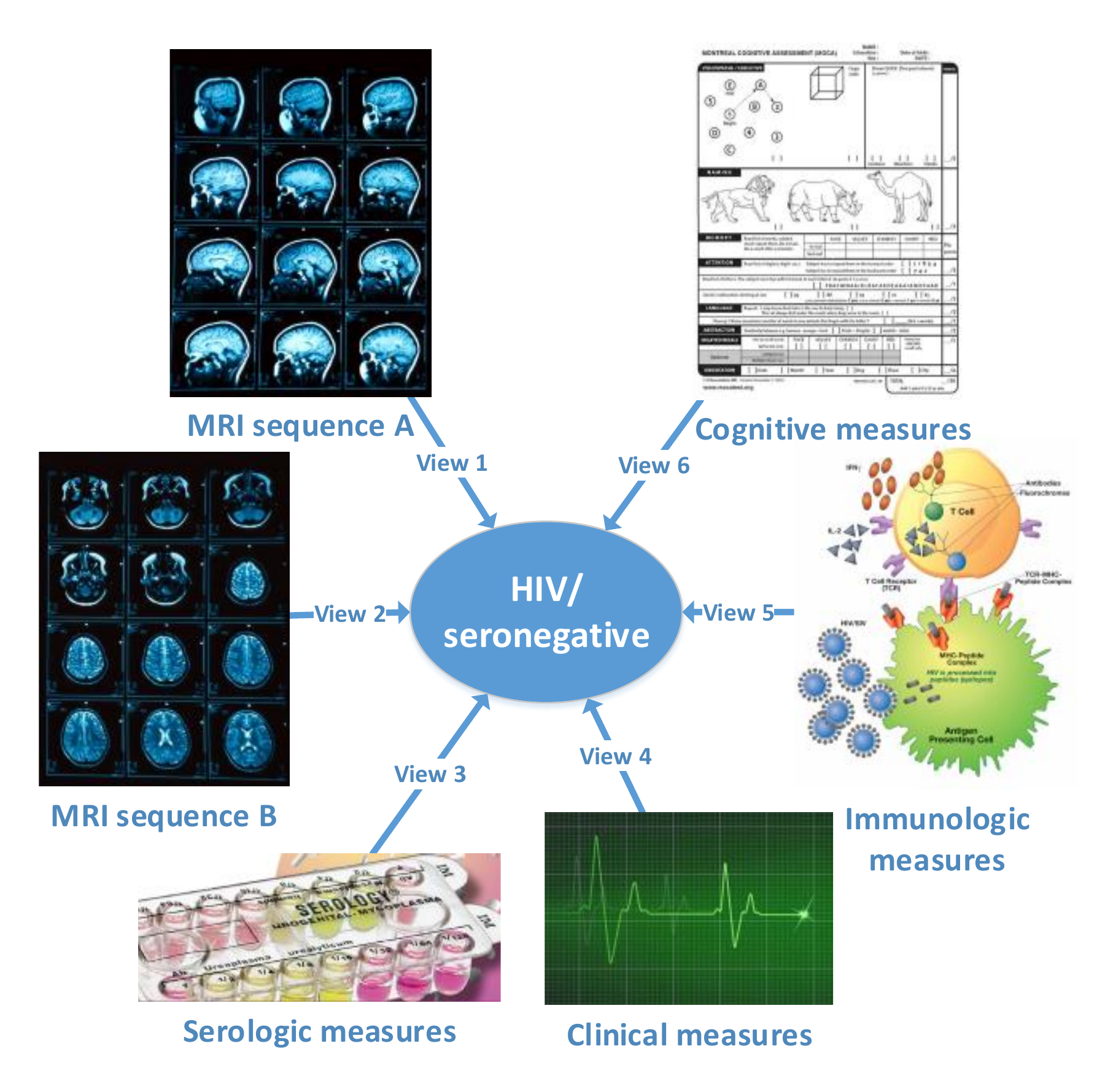}
    \end{minipage}
  \caption{An example of multi-view learning in medical studies \cite{cao2014tensor}.}\label{fig:multiview}
\end{figure}

Medical science witnesses everyday measurements from a series of medical examinations documented for each subject, including clinical, imaging, immunologic, serologic and cognitive measures \cite{cao2015determinants}, as shown in Figure~\ref{fig:multiview}. Each group of measures characterize the health state of a subject from different aspects. This type of data is named as \emph{multi-view data}, and each group of measures form a distinct view quantifying subjects in one specific feature space. Therefore, it is critical to combine them to improve the learning performance, while simply concatenating features from all views and transforming a multi-view data into a single-view data, as the method ($a$) shown in Figure~\ref{fig:tmfs}, would fail to leverage the underlying correlations between different views.

\subsection{Multi-view Learning}

Suppose we have a multi-view classification task with $n$ labeled instances represented from $m$ different views: $\mathcal{D}=\left\{  \left (\mathbf{x}_i^{(1)}, \mathbf{x}_i^{(2)}, \cdots, \mathbf{x}_i^{(m)}, y_i \right)\right\}_{i=1}^n$, where $\mathbf{x}_{i}^{(v)} \in \mathbb{R}^{I_{v} }$, $I_{v}$ is the dimensionality of the $v$-th view, and $y_{i}\in\{-1,+1\}$ is the class label of the $i$-th instance. 

Representative methods for multi-view learning can be categorized into three groups: co-training, multiple kernel learning, and subspace learning \cite{xu2013survey}. Generally, the co-training style algorithm is a classic approach for semi-supervised learning, which trains in alternation to maximize the mutual agreement on different views. Multiple kernel learning algorithms combine kernels that naturally correspond to different views, either linearly \cite{lanckriet2004learning} or nonlinearly \cite{varma2009more,cortes2009learning} to improve learning performance. Subspace learning algorithms learn a latent subspace, from which multiple views are generated. Multiple kernel learning and subspace learning are generalized as co-regularization style algorithms \cite{sun2013survey}, where the disagreement between the functions of different views is taken as a part of the objective function to be minimized. Overall, by exploring the consistency and complementary properties of different views, multi-view learning is more effective than single-view learning.

In the multi-view setting for brain disorders, or for medical studies in general, a critical problem is that there may be limited subjects available (\emph{i.e.}, a small $n$) yet introducing a large number of measurements (\emph{i.e.}, a large $\sum_{i=1}^m I_i$). Within the multi-view data, not all features in different views are relevant to the learning task, and some irrelevant features may introduce unexpected noise. The irrelevant information can even be exaggerated after view combinations thereby degrading performance. Therefore, it is necessary to take care of feature selection in the learning process. Feature selection results can also be used by researchers to find biomarkers for brain diseases. Such biomarkers are clinically imperative for detecting injury to the brain in the earliest stage before it is irreversible. Valid biomarkers can be used to aid diagnosis, monitor disease progression and evaluate effects of intervention \cite{kong2013discriminative}.

Conventional feature selection approaches can be divided into three main directions: filter, wrapper, and embedded methods \cite{guyon2003introduction}. Filter methods compute a discrimination score of each feature independently of the other features based on the correlation between the feature and the label, \emph{e.g.}, information gain, Gini index, Relief \cite{peng2005feature,robnik2003theoretical}. Wrapper methods measure the usefulness of feature subsets according to their predictive power, optimizing the subsequent induction procedure that uses the respective subset for classification \cite{guyon2002gene,rakotomamonjy2003variable,shieh2008multiclass,maldonado2009wrapper,cao2014tensor}. Embedded methods perform feature selection in the process of model training based on sparsity regularization \cite{feng2012adaptive,fang2013discriminative,wang2013multi,wang2013heterogeneous}. For example, Miranda et al. add a regularization term that penalizes the size of the selected feature subset to the standard cost function of SVM, thereby optimizing the new objective function to conduct feature selection \cite{miranda2005linear}. Essentially, the process of feature selection and learning algorithm interact in embedded methods which means the learning part and the feature selection part can not be separated, while wrapper methods utilize the learning algorithm as a black box.

However, directly applying these feature selection approaches to each separate view would fail to leverage multi-view correlations. By taking into account the latent interactions among views and the redundancy triggered by multiple views, it is desirable to combine multi-view data in a principled manner and perform feature selection to obtain consensus and discriminative low dimensional feature representations.


\subsection{Modeling View Correlations}\label{sec:vector_view}

Recent years have witnessed many research efforts devoted to the integration of feature selection and multi-view learning. Tang et al. study multi-view feature selection in the unsupervised setting by constraining that similar data instances from each view should have similar pseudo-class labels \cite{tang2013unsupervised}. Considering brain disorder identification, different neuroimaging features may capture different but complementary characteristics of the data. For example, the voxel-based tensor features convey the global information, while the ROI-based Automated Anatomical Labeling (AAL) \cite{tzourio2002automated} features summarize the local information from multiple representative brain regions. Incorporating these data and additional non-imaging data sources can potentially improve the prediction. For Alzheimer's disease (AD) classification, Ye et al. propose a kernel-based method for integrating heterogeneous data, including tensor and AAL features from MRI images, demographic information and genetic information \cite{ye2008heterogeneous}. The kernel framework is further extended for selecting features (biomarkers) from heterogeneous data sources that play more significant roles than others in AD diagnosis.

Huang et al. propose a sparse composite linear discriminant analysis model for identification of disease-related brain regions of AD from multiple data sources \cite{huang2011identifying}. Two sets of parameters are learned: one represents the common information shared by all the data sources about a feature, and the other represents the specific information only captured by a particular data source about the feature. Experiments are conducted on the PET and MRI data which measure structural and functional aspects, respectively, of the same AD pathology. However, the proposed approach requires the input as the same set of variables from multiple data sources. Xiang et al. investigate multi-source incomplete data for AD and introduce a unified feature learning model to handle block-wise missing data which achieves simultaneous feature-level and source-level selection \cite{xiang2013multi}.

For modeling view correlations, in general, a coefficient is assigned for each view, either at the view-level or feature-level. For example, in multiple kernel learning, a kernel is constructed from each view and a set of kernel coefficients are learned to obtain an optimal combined kernel matrix. These approaches, however, fail to explicitly consider correlations between features.

\subsection{Modeling Feature Correlations}\label{sec:vector_feature}

One of the key issues for multi-view classification is to choose an appropriate tool to model features and their correlations hidden in multiple views, since this directly determines how information will be used. In contrast to modeling on views, another direction for modeling multi-view data is to directly consider the correlations between features from multiple views. Since taking the tensor product of their respective feature spaces corresponds to the interaction of features from multiple views, the concept of tensor serves as a backbone for incorporating multi-view features into a consensus representation by means of tensor product, where the complex multiple relationships among views are embedded within the tensor structures. By mining structural information contained in the tensor, knowledge of multi-view features can be extracted and used to establish a predictive model.

\begin{figure}[t]
\centering
    \begin{minipage}[l]{\columnwidth}
      \centering
      \includegraphics[width=1\textwidth]{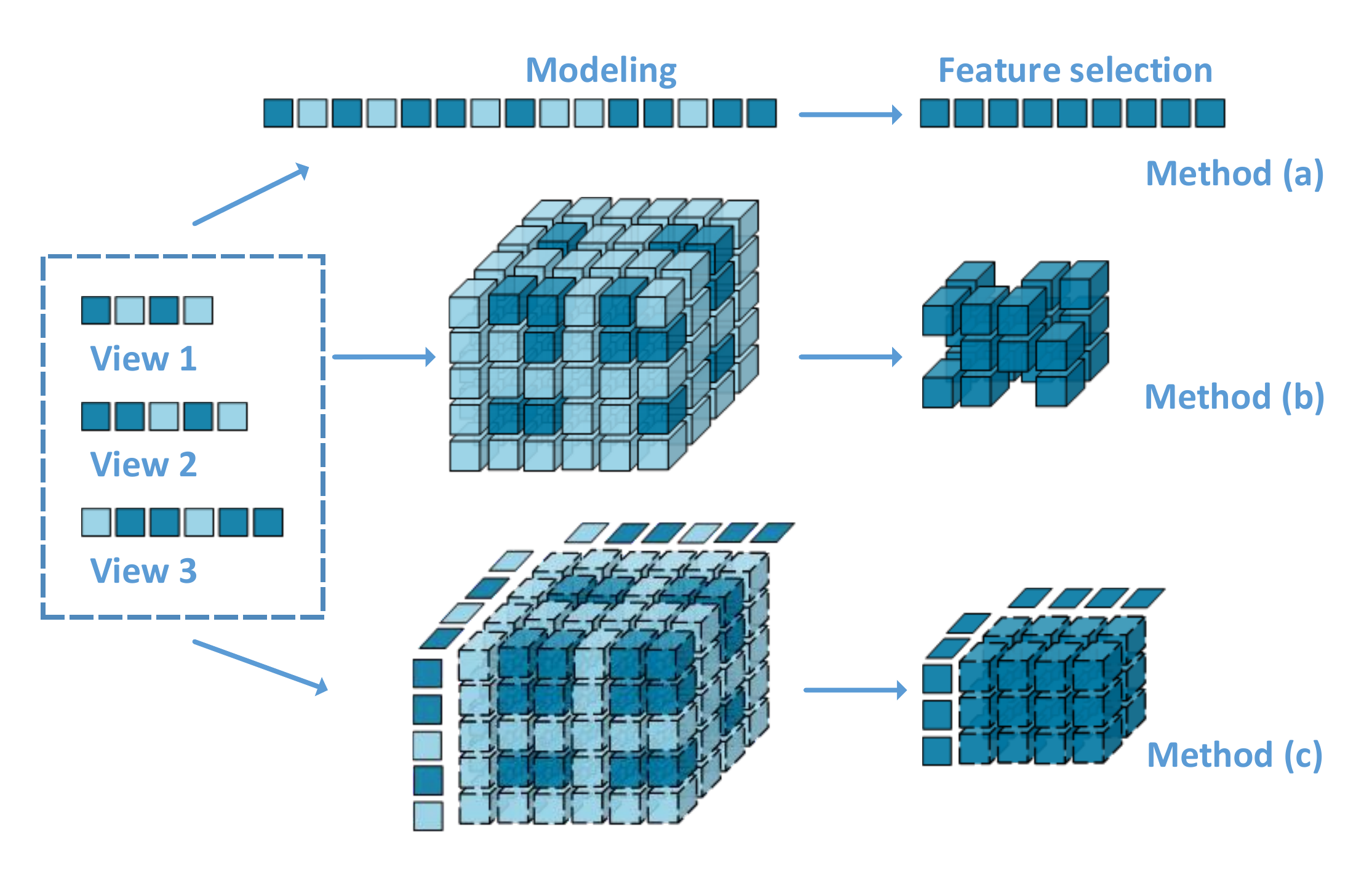}
    \end{minipage}
  \caption{Schematic view of the key differences among three strategies of multi-view feature selection \cite{cao2014tensor}.}\label{fig:tmfs}
\end{figure}

Smalter et al. formulate the problem of feature selection in the tensor product space as an integer quadratic programming problem \cite{smalter2009feature}. However, this method is computationally intractable on many views, since it directly selects features in the tensor product space resulting in the curse of dimensionality, as the method ($b$) shown in Figure~\ref{fig:tmfs}. Cao et al. propose to use a tensor-based approach to model features and their correlations hidden in the original multi-view data \cite{cao2014tensor}. The operation of tensor product can be used to bring $m$-view feature vectors of each instance together, leading to a tensorial representation for common structure across multiple views, and allowing us to adequately diffuse relationships and encode information among multi-view features. In this manner, the multi-view classification task is essentially transformed from an independent domain of each view to a consensus domain as a tensor classification problem.

By using $\mathcal{X}_{i} $ to denote $\prod_{v=1}^{m} \otimes \mathbf{x}_{i}^{(v)}$, the dataset of labeled multi-view instances can be represented as $\mathcal{D}=\{(\mathcal{X}_i,y_i)\}_{i=1}^n$. Note that each multi-view instance $\mathcal{X}_{i}$ is an $m$th-order tensor that lies in the tensor product space $\mathbb{R}^{I_{1} \times \cdots \times I_{m}}$. Based on the definitions of inner product and tensor norm, multi-view classification can be formulated as a global convex optimization problem in the framework of supervised tensor learning \cite{tao2007supervised}. This model is named as \emph{multi-view SVM} \cite{cao2014tensor}, and it can be solved with the use of optimization techniques developed for SVM.


Furthermore, a dual method for multi-view feature selection is proposed in \cite{cao2014tensor} that leverages the relationship between original multi-view features and reconstructed tensor product features to facilitate the implementation of feature selection, as the method ($c$) in Figure~\ref{fig:tmfs}. It is a wrapper model which selects useful features in conjunction with the classifier and simultaneously exploits the correlations among multiple views. Following the idea of SVM-based recursive feature elimination \cite{guyon2002gene}, multi-view feature selection is consistently formulated and implemented in the framework of \emph{multi-view SVM}. This idea can extend to include lower order feature interactions and to employ a variety of loss functions for classification or regression \cite{cao2015multi}.


\section{Future Work}

The human brain is one of the most complicated biological structures in the known universe. While it is very challenging to understand how it works, especially when disorders and diseases occur, dozens of leading technology firms, academic institutions, scientists, and other key contributors to the field of neuroscience have devoted themselves to this area and made significant improvements in various dimensions\footnote{http://www.whitehouse.gov/BRAIN}. Data mining on brain disorder identification has become an emerging area and a promising research direction.

This paper provides an overview of data mining approaches with applications to brain disorder identification which have attracted increasing attention in both data mining and neuroscience communities in recent years. A taxonomy is built based upon data representations, \emph{i.e.}, tensor imaging data, brain network data and multi-view data, following which the relationships between different data mining algorithms and different neuroimaging applications are summarized. We briefly present some potential topics of interest in the future.

\textbf{Bridging heterogeneous data representations.} As introduced in this paper, we can usually derive data from neuroimaging experiments in three representations, including raw tensor imaging data, brain network data and multi-view vector-based data. It is critical to study how to train a model on a mixture of data representations, although it is very challenging to combine data that are represented in tensor space, vector space and graph space, respectively.
There is a straightforward idea of defining different kernels on different feature spaces and combing them through multi-kernel algorithms. However it is usually hard to interpret the results. The concept of side view has been introduced to facilitate the process of mining brain networks, which may also be used to guide supervised tensor learning. It is even more interesting if we can learn on tensors and graphs simultaneously.

\textbf{Integrating multiple neuroimaging modalities.} There are a variety of neuroimaging techniques available characterizing subjects from different perspectives and providing complementary information. For example, DTI contains local microstructural characteristics of water diffusion; structural MRI can be used to delineate brain atrophy; fMRI records BOLD response related to neural activity; PET measures metabolic patterns \cite{wee2012identification}. Based on such multimodality representation, it is desirable to find useful patterns with rich semantics. For example, it is important to know which connectivity between brain regions is significant in the sense of both structure and functionality. On the other hand, by leveraging the complementary information embedded in the multimodality representation, better performance on disease diagnosis can be expected.

\begin{figure}[t]
\centering
    \begin{minipage}[l]{\columnwidth}
      \centering
      \includegraphics[width=1\textwidth]{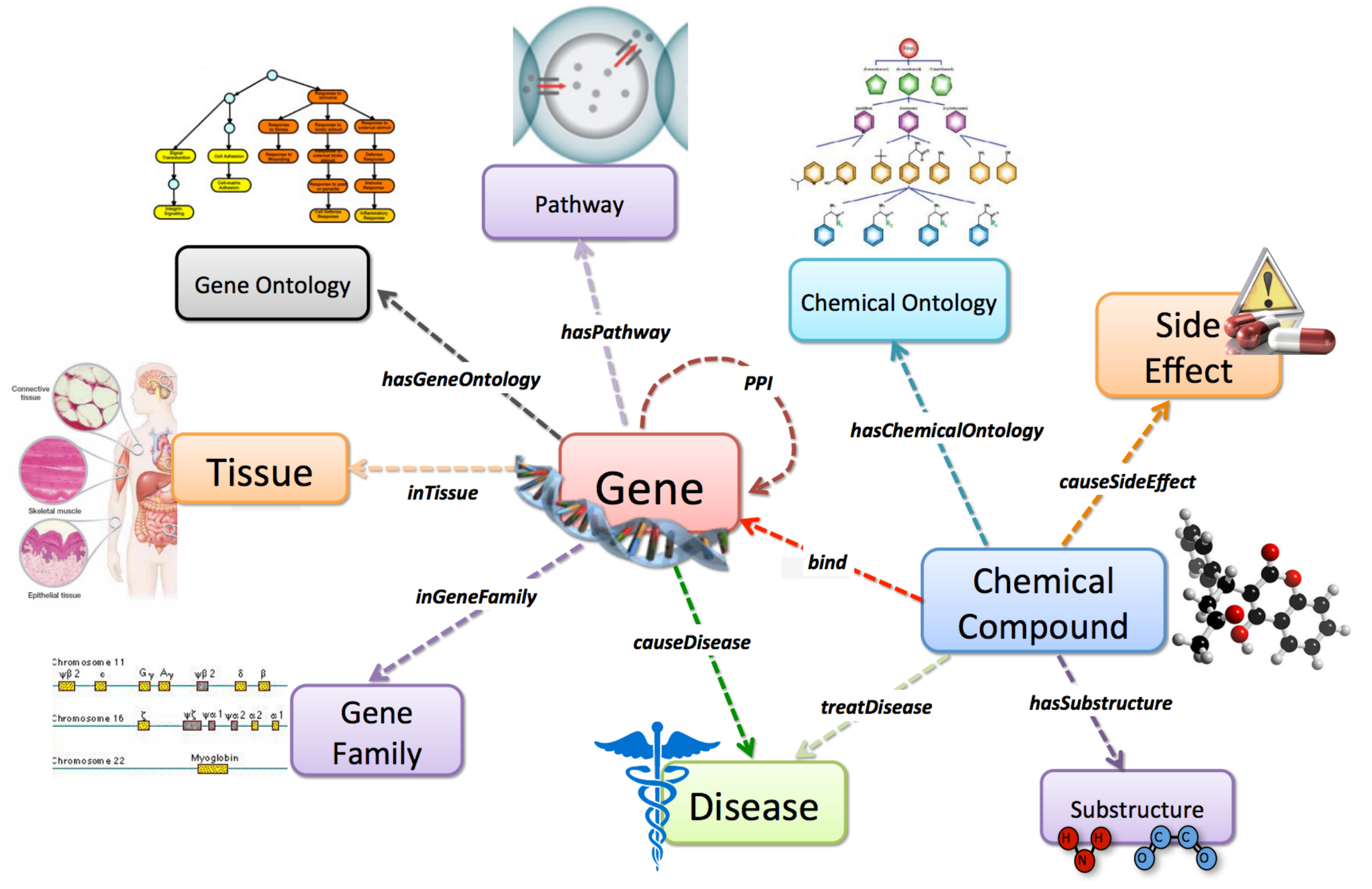}
    \end{minipage}
  \caption{A bioinformatics heterogeneous information network schema.}\label{fig:slap}
\end{figure}

\textbf{Mining bioinformatics information networks.} Bioinformatics network is a rich source of heterogeneous information involving disease mechanisms, as shown in Figure~\ref{fig:slap}. The problems of gene-disease association and drug-target binding prediction have been studied in the setting of heterogeneous information networks \cite{cao2014collective,kong2013multi}. For example, in gene-disease association prediction, different gene sequences can lead to certain diseases. Researchers would like to predict the association relationships between genes and diseases. Understanding the correlations between brain disorders and other diseases and the causality between certain genes and brain diseases can be transformative for yielding new insights concerning risk and protective relationships, for clarifying disease mechanisms, for aiding diagnostics and clinical monitoring, for biomarker discovery, for identification of new treatment targets and for evaluating effects of intervention.


\bibliographystyle{abbrv}
\bibliography{reference}

\end{document}